\documentclass[conference]{IEEEtran}
\IEEEoverridecommandlockouts
\usepackage{cite}
\usepackage{amsmath,amssymb,amsfonts}
\usepackage{algorithm}
\usepackage{algpseudocode}
\usepackage{graphicx}
\usepackage{textcomp}
\usepackage{xcolor}
\usepackage{subcaption}
\usepackage{booktabs}
\def\BibTeX{{\rm B\kern-.05em{\sc i\kern-.025em b}\kern-.08em
    T\kern-.1667em\lower.7ex\hbox{E}\kern-.125emX}}
\begin{document}

\title{Rethinking Video Super-Resolution: Towards Diffusion-Based Methods without Motion Alignment}

\author{
\IEEEauthorblockN{
Zhihao Zhan\textsuperscript{1}\textsuperscript{\dag}\thanks{\textsuperscript{\dag} These authors contributed equally.}, 
Wang Pang\textsuperscript{2}\textsuperscript{\dag},
Xiang Zhu\textsuperscript{1}\textsuperscript{\dag} 
and Yechao Bai\textsuperscript{2}\textsuperscript{*}} 

\IEEEauthorblockA{\textsuperscript{1} TopXGun Robotics, Nanjing, 211100, China}

\IEEEauthorblockA{\textsuperscript{2} Nanjing University, Nanjing, 210023, China}

\IEEEauthorblockA{Email: zhzhan@topxgun.com, 201180035@smail.nju.edu.cn, xzhu@topxgun.com, ychbai@nju.edu.cn}
}

\maketitle

\begin{abstract}
In this work, we rethink the approach to video super-resolution by introducing a method based on the Diffusion Posterior Sampling framework, combined with an unconditional video diffusion transformer operating in latent space. The video generation model, a diffusion transformer, functions as a space-time model. We argue that a powerful model, which learns the physics of the real world, can easily handle various kinds of motion patterns as prior knowledge, thus eliminating the need for explicit estimation of optical flows or motion parameters for pixel alignment. Furthermore, a single instance of the proposed video diffusion transformer model can adapt to different sampling conditions without re-training. Empirical results on synthetic and real-world datasets illustrate the feasibility of diffusion-based, alignment-free video super-resolution.
\end{abstract}


\section{Introduction}
The concept of super-resolution was first proposed in the 1980s \cite{tsai1984multiframe, yang2017image}, primarily focusing on multi-frame image super-resolution, also known as video super-resolution (VSR). 
The fundamental principle involves aligning and fusing image information of the same object across multiple frames to surpass the Nyquist limit. 
This process represents a typical inverse problem, requiring sub-pixel spatial alignment across frames, along with resampling and deconvolution to achieve enhanced resolution.

Over the past decade, the primary focus of super-resolution has shifted towards single image super-resolution (SISR), which eliminates the need for spatial alignment or motion estimation. 
The recovery of high-frequency components in SISR predominantly relies on deep neural networks such as convolutional neural networks (CNNs) \cite{dong2015image,tai2017image,zhang2018image}. 
These networks are capable of mapping low-resolution (LR) input image to the corresponding high-resolution (HR) output, mimicking the behavior of deconvolution. 
Such methods are effective when the upscaling factor is less than 4x; however, beyond this value, the output images tend to appear overly smoothed.

Since 2022, diffusion models (DMs) \cite{ho2020denoising, song2020score} have become increasingly important in SISR. 
They facilitate the super-resolution of images with large upscaling factors (e.g., 4x, 8x, 16x) \cite{saharia2022image,chung2022diffusion}. 
This effectiveness arises because DMs can learn the distribution of the underlying HR images as prior knowledge. 
They are capable of synthesizing realistic high-frequency components in their outputs, which often result in sharper images compared to those produced solely by deep neural networks. 
In such instances, human preference, often measured via metrics such as the fool rate \cite{saharia2022image}, becomes the primary metric for evaluating super-resolution algorithms. 
However, the fidelity of their outputs with respect to the underlying HR reference cannot be guaranteed.

Meanwhile, there are still scenarios where high fidelity to the real world is crucial, such as in remote sensing \cite{zhu2020super}, medical imaging \cite{peng2020saint}, and surveillance monitoring \cite{deshmukh2019fractional}. 
In these cases, exploring temporal redundancies across video frames to generate HR images is beneficial. 
Studies have shown that inter-frame information can greatly enhance super-resolution results \cite{liu2022video}. 
Unfortunately, that also means we have to face the motion estimation issue, which is challenging due to the diverse motion patterns present in the real world. 
Recent VSR algorithms are generally categorized into two types: those with explicit motion estimation and those without.

Methods with motion estimation typically incorporate an optical flow component to achieve sub-pixel spatial alignment and warp each frame to compensate for pixel movement \cite{kappeler2016video,li2018video,li2021comisr,zhou2024upscale,youk2024fma}. 
However, these steps often introduce errors, leading to artifacts. 
Zhou et al. utilize RAFT \cite{teed2020raft} to estimate flow and selectively warp pixels with high forward-backward consistency, which helps mitigate estimation errors \cite{zhou2024upscale}. 
In the compensation step, Xu et al. employ a coordinate network designed to minimize resampling artifacts \cite{xu2024enhancing}. 
Nonetheless, the effectiveness of these methods is generally constrained by the capabilities of their motion estimation components, which often struggle to handle complex motion patterns, especially when rotations, occlusions, and non-rigid object behaviors are involved.

Methods that forego motion estimation typically utilize deep networks (e.g., CNNs, RNNs) or a diffusion process to smooth video content across frames, thereby maintaining consistency in their super-resolved outputs over short temporal periods \cite{jo2018deep,kim2019video,fuoli2019efficient,chen2024learning}. 
However, these approaches often cannot capture long-term pixel dependencies, which result in relatively weaker restoration efficiency compared to methods with motion estimation.

An interesting study by Shi et al. demonstrated that transformers \cite{vaswani2017attention} can directly capture subtle movements across frames through their attention mechanism \cite{shi2022rethinking}, although a patch-based alignment step is still required to reduce overall motion magnitude in their method. 
This finding suggests that transformers are particularly well-suited for precise motion estimation in video processing.

In this paper, we introduce a novel VSR algorithm, based on an unconditional video diffusion model (VDM), which fundamentally differs from the deep learning-based approaches previously discussed. 
Unlike most existing VSR methods that generate HR outputs from LR videos through supervised training, our approach handles VSR as solving an inverse problem. 
This involves both conditioning likelihood estimation from LR observations and prior probability estimation of the underlying HR video, akin to the maximum-a-posteriori (MAP) algorithms \cite{protter2008generalizing,yang2010image} used before the emergence of learning-based super-resolution methods.

However, our method diverges from traditional MAP-based VSRs by utilizing powerful unconditional diffusion models as a tool to represent prior knowledge, grounded in the Diffusion Posterior Sampling (DPS) framework \cite{chung2022diffusion}. 
The original DPS has already shown its robust reconstruction capabilities in SISR and blind global motion deblurring \cite{chung2022diffusion,chung2023parallel}. 
Our algorithm introduces several novelties:

\begin{enumerate}
\item It processes 3D videos in their entirety, learning their distribution across both spatial and temporal axes.
\item By integrating a transformer network as the denoiser in the reverse diffusion process, the model achieves superior scalability and enhanced effectiveness in managing complex and dynamic scenarios.
\item The method captures the statistical properties of visual data within a latent space, thereby achieving dimensionality reduction.
\item It verifies that the incorporation of inter-frame motion information can improve the performance of VSR. 
\end{enumerate}

This algorithm is founded on our belief that with unconditional DMs, explicit motion estimation over time is unnecessary, akin to how facial symmetry is naturally maintained in face image generation without explicit interventions \cite{peebles2023scalable}. However, it demands that the DMs not only learn the distribution of video contents but also understand their dynamics governed by real-world physics. To validate our approach, we employ synthetic and real-world data to explore the behavioral dynamics of the method.

\section{Related work}
\label{sec:related}

\subsection{Diffusion models}
Diffusion models have achieved huge success in generating a wide array of multi-dimensional signals, including images, videos, audios, and texts. These models excel at learning the prior distribution of the underlying signals \cite{song2020score}. During training, scheduled Gaussian noise is systematically added to a clean signal sample $\mathbf{x}$ until it is transformed into pure Gaussian noise. Simultaneously, a network is trained to reverse this noising process by learning to predict the noise at each step. In the reverse phase, this trained network begins with a sample of pure Gaussian noise and incrementally denoises it, aiming to generate a signal that faithfully represents the distribution of the training dataset $\{\mathbf{x}\}$.

According to Song et al. \cite{song2020score}, the iterative noising process can be described by the following stochastic differential equation (SDE):
\begin{equation}
  d\mathbf{x} = -\frac{\beta(t)}{2}\mathbf{x}_t dt + \sqrt{\beta(t)}d\mathbf{w},
  \label{eq:sde_forward2}
\end{equation}
where $\beta(t)$ denotes the noise schedule \cite{song2020score}, $\mathbf{w}$ represents a standard Brownian motion, and $d\mathbf{w}$ is considered as white Gaussian noise. 

Given the forward equation \eqref{eq:sde_forward2}, the reverse process that denoises a sampled Gaussian signal back to the data distribution should theoretically be:
\begin{equation}
  d\mathbf{x} = \left(-\frac{\beta(t)}{2}\mathbf{x}_t - \beta\nabla_{\mathbf{x}_t} \log p(\mathbf{x}_t) \right)dt + \sqrt{\beta(t)}d\mathbf{w},
  \label{eq:sde_reverse1}
\end{equation}
where $\nabla_{\mathbf{x}_t} \log p(\mathbf{x}_t)$ is the score function of the unknown distribution $p(\mathbf{x}_t)$. This score function can be approximated by a neural network $\mathbf{s}_{\theta}(\mathbf{x}_{t}, t)$ via score matching:
\begin{equation}
  \theta^{*} = \arg\min_{\theta}\mathbb{E}_{t, \mathbf{x}_t, \mathbf{x}_0}\left( \| \nabla_{\mathbf{x}_t} \log p(\mathbf{x}_t|\mathbf{x}_0) - \mathbf{s}_{\theta}(\mathbf{x}_{t}, t) \|_{2}^{2} \right),
  \label{eq:score_matching}
\end{equation}
where $\mathbf{s}_{\theta}(\mathbf{x}_{t}, t)$ is time-dependent and can replace the score function $\nabla_{\mathbf{x}_t} \log p(\mathbf{x}_t)$ in \eqref{eq:sde_reverse1}. 

In many diffusion applications, additional conditions are imposed on the score function to guide and control the outputs. For instance, text prompts can be used to specify the contents of the generated images or videos, or LR images can serve as guides for generating HR images. Typically, these conditional signals are represented as a vector $\mathbf{c}$ and directly incorporated into the network. This integration enables the conditional diffusion process to be mathematically expressed as follows:
\begin{equation}
  \nabla_{\mathbf{x}_t} \log p(\mathbf{x}_t | \mathbf{c}) \approx 
  \mathbf{s}_{\theta}(\mathbf{x}_t, t, \mathbf{c}).
  \label{eq:conditional_score}
\end{equation}

\subsection{Diffusion Posterior Sampling (DPS)}
Chung et al. addressed challenges such as image deblurring, inpainting, and SISR by treating them as inverse problems and proposed the DPS framework as a generic solver for such problems \cite{chung2022diffusion}.

In an inverse problem, suppose $\mathbf{x}$ represents an ideal data vector to be estimated, and $\mathbf{y}$ denotes the observation of $\mathbf{x}$ in the real world. the observation $\mathbf{y}$ usually has a lower dimension than $\mathbf{x}$, making the recovery of $\mathbf{x}$ from $\mathbf{y}$ ill posed. The degradation model that transforms $\mathbf{x}$ to $\mathbf{y}$ is assumed to be known, hence we have the conditional probability function $p(\mathbf{y} | \mathbf{x})$. For example, if the degradation model is:
\begin{equation}
  \mathbf{y} = H(\mathbf{x}) + \mathbf{e},
  \label{eq:observation}
\end{equation}
where $\mathbf{e}$ denotes Gaussian sensing noise with standard deviation $\sigma$, the conditional probability can be written as:
\begin{equation}
  p(\mathbf{y} | \mathbf{x}) = \mathcal{N}(\mathbf{y} | H(\mathbf{x}), \sigma^{2}\mathbf{I} ).
  \label{eq:conditional_prob}
\end{equation}
With the conditional and the prior probabilities, according to the Bayesian rule, we can derive the corresponding conditional score function as
\begin{equation}
  \nabla_{\mathbf{x}} \log p(\mathbf{x} | \mathbf{y}) = 
  \nabla_{\mathbf{x}} \log p(\mathbf{y} | \mathbf{x}) + \nabla_{\mathbf{x}} \log p(\mathbf{x}).
  \label{eq:posterior_score}
\end{equation}

In each iteration of the denoising process, Chung et al. used the following approximation to update the conditional likelihood \cite{chung2022diffusion}:
\begin{equation}
  \nabla_{\mathbf{x}_t} \log p(\mathbf{y} | \mathbf{x}_t) \approx 
  \nabla_{\mathbf{x}_t} \log p(\mathbf{y} | \hat{\mathbf{x}}_0(\mathbf{x}_t)),
  \label{eq:dps_approx1}
\end{equation}
where
\begin{equation}
  \hat{\mathbf{x}}_0(\mathbf{x}_t) = \frac{1}{\sqrt{\bar{\alpha}(t)}}(\mathbf{x}_t + (1 - \bar{\alpha}(t))\mathbf{s}_{\theta^*}(\mathbf{x}_t, t)).
  \label{eq:dps_approx2}
\end{equation}

Combining the conditional probability model \eqref{eq:conditional_prob}, we can rewrite the reverse iteration function \eqref{eq:sde_reverse1} as:
\begin{equation}
\begin{split}
  d\mathbf{x} = & \biggl[-\frac{\beta(t)}{2}\mathbf{x}_t - \beta\Bigl( \mathbf{s}_{\theta^*}(\mathbf{x}_t, t) \\
  & - \frac{1}{\sigma^2} \nabla_{\mathbf{x}_t} \|\mathbf{y} - H(\hat{\mathbf{x}}_0(\mathbf{x}_t)) \| \Bigr) \biggr] dt + \sqrt{\beta(t)}d\mathbf{w}.
  \label{eq:dps_reverse}
\end{split}
\end{equation}
Note that, compared with the conditional DM solution \eqref{eq:conditional_score}, the above DPS method utilizes an unconditional network $\mathbf{s}_{\theta^*}(\mathbf{x}_t, t)$, and hence once the network is trained, it can be used across different inverse problems and under various sensing model settings. 

\subsection{Video Diffusion Models (VDMs)}
Recently, some video DMs have demonstrated highly impressive results in realistic video generation \cite{gupta2023photorealistic, ma2024latte, openai2024sora, opensora}. Most of these models utilize transformer-based network architectures, renowned for their strong scalability and parallelization capabilities. Several networks are essentially extensions of the DiT image generation model \cite{peebles2023scalable}.

A notable example is the Sora model released by OpenAI in 2024 \cite{openai2024sora}, which produced realistic results that captivated the global audience. Analysis of videos generated by Sora reveals two important characteristics: 
\begin{enumerate}
\item Strong temporal coherence across frames;
\item Realistic object movement simulations that emulate the physics of the real world.
\end{enumerate}
 
Furthermore, once the VDM has learned the underlying dynamics of a world represented by a training video dataset, it can naturally resolve single image motion blur affected by intra-frame motion as long as the image is about the given world \cite{pang2025imagemotionblurremoval}. 

These observations prompt us to consider the following question: Given that these video diffusion models can consistently track objects with a variety of complex movements, effectively mirroring real-world dynamics, can VDMs be effectively applied to VSR within a straightforward inverse problem-solving framework?

\section{Proposed Approach}
\label{sec:approach}
\begin{figure*}
    \centering
    \includegraphics[width=0.9\linewidth]{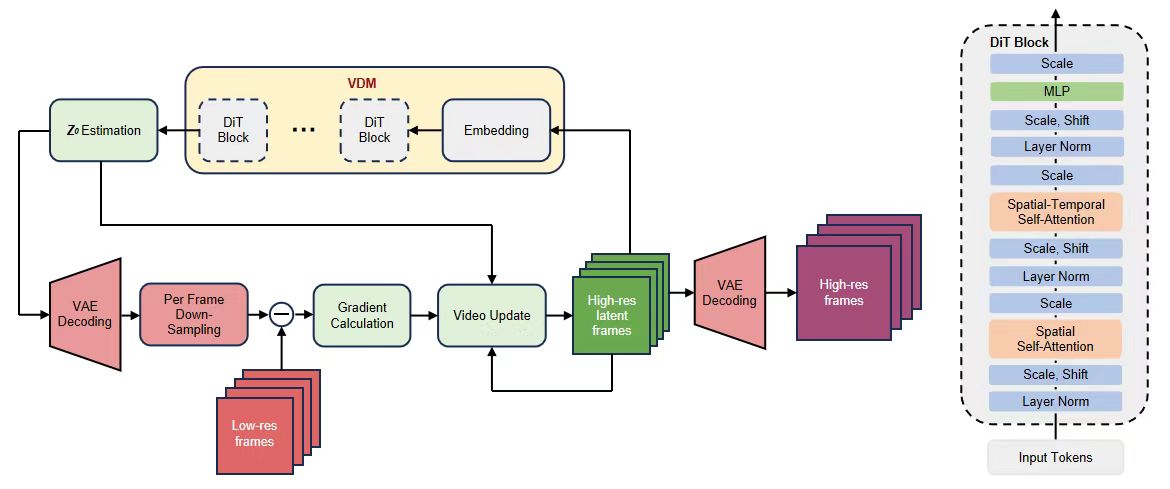}
    \caption{Overview of the Video Diffusion Model based VSR (VDM-VSR): In the core iteration, the estimated 3D HR video resides in the latent space, represented by green boxes. It is generated and refined by the VDM, which includes several Transformer blocks, as shown in the structural diagram on the right. The latent video is then decoded and compared with the LR observations through the degradation model, indicated by red boxes. The discrepancies between these observations and the latent video are used to correct and enhance the HR video during the iteration. Upon completion of this iteration, the latent video is decoded back to the conventional HR space. }
    \label{fig:overview}
    \vspace{-1.8em} 
\end{figure*}

We introduce VDM-VSR, an approach that addresses super-resolution as an inverse problem under the DPS framework. This method utilizes a DiT-based VDM as the denoiser. We posit that as long as the VDM effectively learns the dynamics of the world as represented by the training video dataset, it should inherently manage inter-frame motion estimation. An overview of the proposed algorithm's architecture is depicted in Fig.~\ref{fig:overview}. 

Let us first examine the degradation model employed by our method. Assume an observed LR video $\mathbf{Y}$ consists of $f$ frames, denoted as $\mathbf{Y} = \{\mathbf{y}_{1:f}\} \in \mathbb{R}^{f \times h \times w \times 3}$. The degradation of each frame can be modeled by the equation:
\begin{equation}
  \mathbf{y}_j = L(\mathbf{x}_j \ast \mathbf{h}) + \mathbf{e},
  \label{eq:frame_degrade}
\end{equation}
where $\mathbf{x}_j$ represents the $j$th HR frame, $\mathbf{h}$ is a known blur kernel, $\ast$ denotes the convolution operator, and $L(\cdot)$ is the down-sampling function. This model employs spatial down-sampling to degrade the spatial dimension, resulting in visual blur. We assume that the sensing noise $\mathbf{e}$ is white Gaussian with a covariance matrix $\sigma^2\mathbf{I}$.

In subsequent sections of the paper, we will use $\mathbf{X} \in \mathbb{R}^{F \times H \times W \times 3}$ to denote the HR frame sequence and simplify the degradation model to:
\begin{equation}
\mathbf{Y} = H(\mathbf{X}, \mathbf{h}) + \mathbf{E}.
\label{eq:frame_degrade_2}
\end{equation}
where $H(\mathbf{X}, \mathbf{h})$ is a simplified formulation of the frame-wise degradation process, which approximates the combined effects of spatial convolution and downsampling applied to each frame $\mathbf{x}_j$ in the sequence.

Similar to DPS, the prior of $\mathbf{Y}$ is managed by a trained diffusion model. However, unlike the original DPS, our diffusion sampling occurs in a low-dimensional latent space defined by a pre-trained variational autoencoder (VAE) \cite{kingma2013auto, rombach2022high}. This adaptation is necessary due to the large dimensionality of video data, which requires reduction to conserve computational resources. The VAE is applied only in the spatial domain and the compression factor $p=8$, resulting in the representation of $\mathbf{X}$ in the VAE latent space being denoted as $\mathbf{Z} \in \mathbb{R}^{F \times H/p \times W/p \times c}$, where $c$ represents the VAE channel size. $\mathbf{X}$ can be reconstructed from $\mathbf{Z}$ using the corresponding VAE decoder $D(\cdot)$.

Given an estimated latent HR video $\mathbf{Z}$, the conditional probability of the observed $\mathbf{Y}$ is expressed as:
\begin{equation}
  p(\mathbf{Y} | \mathbf{Z}) = \mathcal{N}(\mathbf{Y} | H(D(\mathbf{Z}), \mathbf{h}), \sigma^2 \mathbf{I}).
  \label{eq:new_conditional_prob}
\end{equation}

Note that $H(D(\cdot), \mathbf{h})$, while no longer linear, remains differentiable and, as such, can be integrated into the DPS framework. The corresponding reverse iteration function is formulated as:
\begin{equation}
\begin{split}
  d\mathbf{Z} = & \biggl[-\frac{\beta(t)}{2}\mathbf{Z}_t - \beta\Bigl( \mathbf{s}_{\theta^*}(\mathbf{Z}_t, t) \\
  & - \frac{1}{\sigma^2} \nabla_{\mathbf{Z}_t} \|\mathbf{Y} - H(D(\hat{\mathbf{Z}}_0(\mathbf{Z}_t)), \mathbf{h}) \| \Bigr) \biggr] dt \\
  & + \sqrt{\beta(t)}d\mathbf{W}.
  \label{eq:our_dps_reverse}
\end{split}
\end{equation}

The unconditional diffusion model $\mathbf{s}_{\theta^*}(\mathbf{Z}, t)$ is pre-trained in the latent video space. We utilize a DiT-based neural network for this purpose, which is structurally similar to the STDiT model from the OpenSora project \cite{opensora}, but it excludes any conditional embedding components. The detailed steps of the overall reverse process are outlined in Algorithm~\ref{alg:main}.

\begin{algorithm}
\caption{VDM-VSR}\label{alg:main}
\begin{algorithmic}[1]
\Require $\mathbf{Y}$, $T$, $\mathbf{h}$
\State $\mathbf{Z}_T \sim \mathcal{N}(\mathbf{0}, \mathbf{I})$
\For{$t = T - 1$ to $0$}
\State $\hat{\mathbf{s}} = \mathbf{s}_{\theta^*}(\mathbf{Z}_t, t)$
\State $\hat{\mathbf{Z}}_0 = \frac{1}{\sqrt{\bar{\alpha}_t}} (\mathbf{Z}_t + (1 - \bar{\alpha}_t)\hat{\mathbf{s}})$
\State $\mathbf{\epsilon} \sim \mathcal{N}(\mathbf{0}, \mathbf{I})$
\State $ \mathbf{Z}^{\prime}_{t-1} = \frac{\sqrt{\alpha_t} (1 - \bar{\alpha}_{t-1})}{1 - \bar{\alpha}_t}\mathbf{Z}_t + \frac{\sqrt{\bar{\alpha}_{t-1}}\beta_t}{1 - \bar{\alpha}_t}\hat{\mathbf{Z}}_0 + \sigma_t\mathbf{\epsilon}$
\State $\hat{\mathbf{Y}}_{t-1} = H(D(\hat{\mathbf{Z}}_0), \mathbf{h})$
\State $ \mathbf{Z}_{t-1} = \mathbf{Z}^{\prime}_{t-1} - \eta_t \nabla_{\mathbf{Z}_t}\|\mathbf{Y} - \hat{\mathbf{Y}}_{t-1} \|_2^2 $
\EndFor
\State $\hat{\mathbf{X}} = D(\hat{\mathbf{Z}}_0)$
\State \Return{$\hat{\mathbf{X}}$}
\end{algorithmic}
\vspace{-0.2em} 
\end{algorithm}

\section{Experiments}
\label{sec:experiment}

\subsection{Synthetic Dataset}
To analyze our algorithm’s behavior without requiring a large-scale transformer model and a huge video dataset, we utilized the synthetic Moving MNIST dataset \cite{srivastava2015unsupervised} as a representation of a 'toy world'. This dataset features limited types of content, with movements governed by simple physics rules. We trained a Moving MNIST VDM with around 20k videos, each video contains 10 frames.

Initially, we aimed to verify that inter-frame information enhances the results of this algorithm. We incorporated an additional frame masking process, $M(\cdot)$, into the degradation model \eqref{eq:frame_degrade_2}, allowing only selected frames to contribute to the restored HR video. 
\begin{equation}
\mathbf{Y} = M(H(\mathbf{X}, \mathbf{h})) + \mathbf{E}.
\label{eq:frame_degrade_3}
\end{equation}
Note that the number of frames in the underlying HR video remains unchanged. 

\begin{figure}
    \centering
    \includegraphics[width=0.65\linewidth]{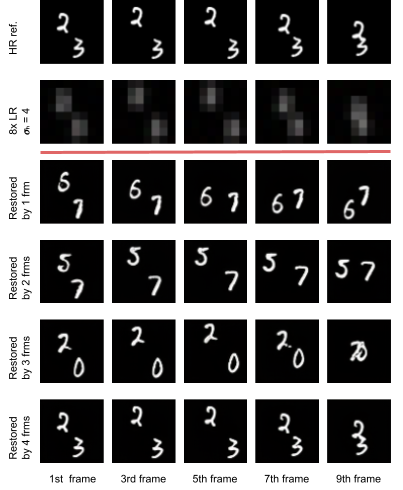}
    \caption{A 64x64x10 Moving MNIST frame sequence, its 8x down-sampled version, and the super-resolved results using different number of frames.}
    \label{fig:moving_mnist}
    \vspace{-1em} 
\end{figure}

\begin{figure}
    \centering
    \includegraphics[width=0.85\linewidth]{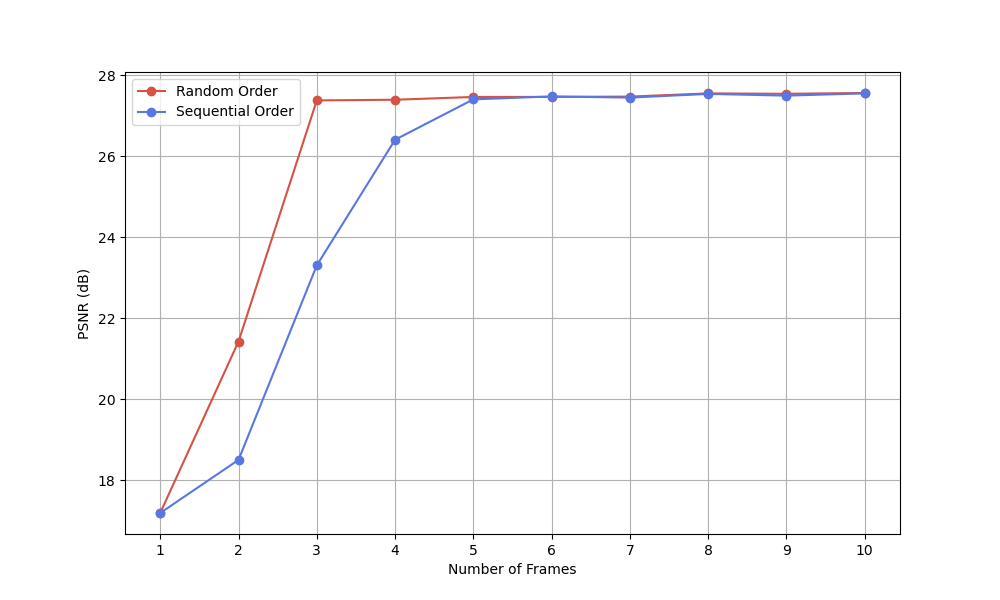}
    \caption{PSNR v.s. number of used frames. The number of input frames gradually increases, and the PSNRs of the 1st frame are recorded. Each PSNR value in this plot is an average over 8 reference videos, and each video are restored 10 times with different noise instances.}
    \label{fig:psnr_plot_1}
    \vspace{-1em} 
\end{figure}

In the first experiment, we set the down-sampling factor to 8x and progressively increased the number of frames used from 1 to 10. We selected 8 HR reference videos and used 10 different noise instances for each to analyze the algorithm's average behavior. An example using one noise instance is shown in Fig.~\ref{fig:moving_mnist}, and the averaged PSNR for each frame number is plotted in the blue line of Fig.~\ref{fig:psnr_plot_1}.

From Fig.~\ref{fig:moving_mnist}, it is apparent that when the observed frame number is low, both video content and motion over frames are incorrectly estimated. Similarly, the plot in Fig.~\ref{fig:psnr_plot_1} shows that the PSNR steadily increases from frame number 1 to 5, indicating that pixel information from subsequent frames aids the restoration of the first frame. However, beyond 5 frames, the PSNR value plateaus. Visually, the corresponding outputs appear almost identical to the reference video, indicating that the restoration quality has reached saturation.

In the second experiment, frames were added incrementally but in a random order. When comparing its PSNR performance (see the red line in Fig.~\ref{fig:psnr_plot_1}) to the sequential order, it is evident that the PSNR generally increases more rapidly with the random sequence. This suggests:
\begin{enumerate}
\item The algorithm implements a global multi-frame super-resolution approach along the frame/time axis, rather than merely smoothing neighboring frames.
\item The random frame order outperforms the sequential one because it better captures object motion with frames that are more widely spaced on the time axis.
\end{enumerate}

Furthermore, our purpose is to address a dilemma concerning the sampling of LR images in super-resolution. To preserve more high-frequency components in the super-resolved image, it is preferable to have more aliasing in the LR frames. Conversely, for accurate spatial alignment of these frames, reducing aliasing is essential. This dilemma has significantly restricted the performance of existing VSR methods.

\begin{figure}
    \centering
    \includegraphics[width=0.65\linewidth]{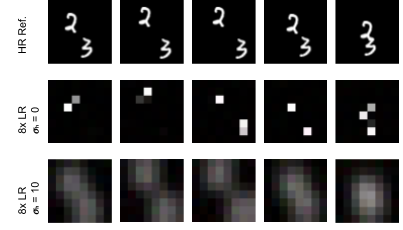}
    \caption{A Moving MNIST frame sequence and its 8x down-sampled versions with blur kernel $\sigma_h = 0$ and $10$ pixels.}
    \label{fig:aliasing_mnist}
    \vspace{-1em} 
\end{figure}

\begin{figure}
    \centering
    \begin{subfigure}[b]{0.5\textwidth}
        \centering
        \includegraphics[width=0.85\linewidth]{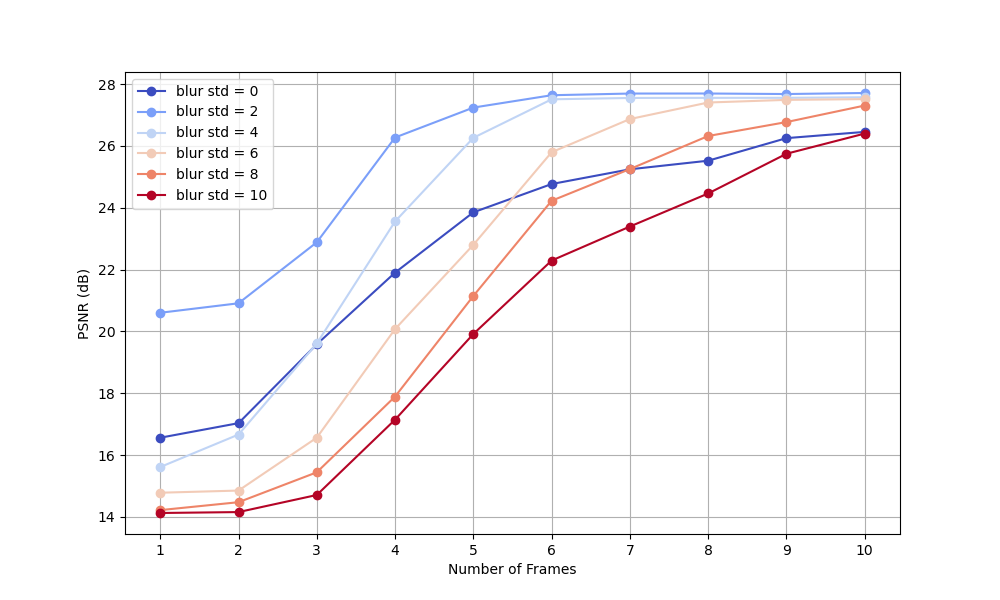}
        \caption{4x super-resolution}
    \end{subfigure}
    
    \begin{subfigure}[b]{0.5\textwidth}
        \centering
        \includegraphics[width=0.85\linewidth]{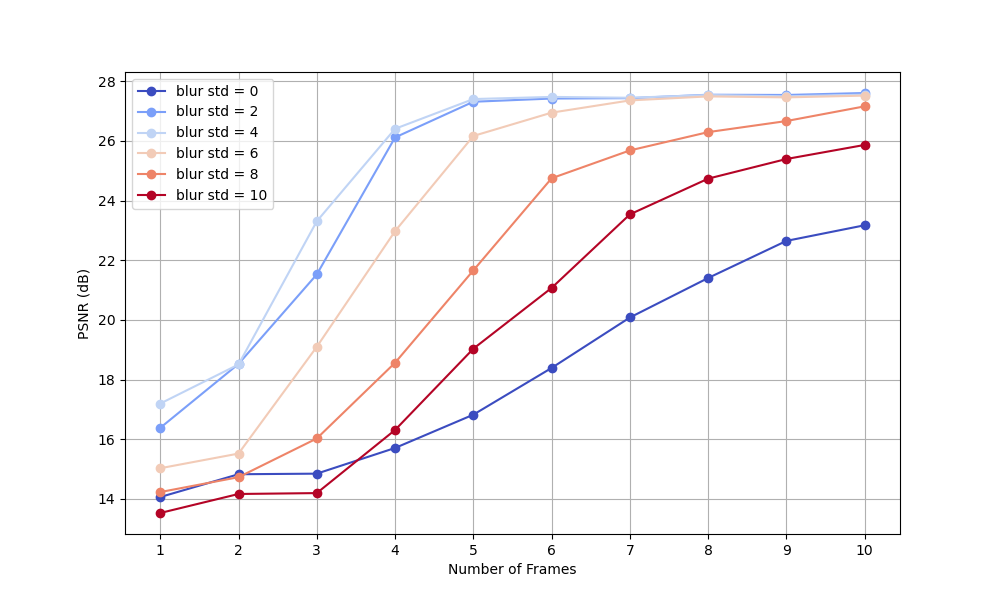}
        \caption{8x super-resolution}
    \end{subfigure}
    
    \caption{PSNR v.s. number of used frames. LR inputs are generated with blur kernel $\sigma_h = 0, 2, 4, 6, 8, 10$.}
    \label{fig:psnr_plot_2}
    \vspace{-1em} 
\end{figure}

To examine how aliasing influences the behavior of our proposed algorithm, we adjusted the blur kernel $\mathbf{h}$ in \eqref{eq:frame_degrade_3}, modeled with a Gaussian shape. We aimed to observe changes in PSNR versus the kernel's spread (represented by its standard deviation $\sigma_h$) and the number of used frames. Highly aliased videos and overly smoothed videos are tested (see examples in Fig.~\ref{fig:aliasing_mnist}). The results, presented in Fig.~\ref{fig:psnr_plot_2}, show that the PSNR trajectories for all values of $\sigma_h$ converge to approximately the same value as the number of frames increases. This finding suggests that regardless of the level of aliasing in the input LR frames, the algorithm can achieve the optimal solution provided a sufficient number of frames are available. Thus, the trade-off between aliasing and spatial accuracy can be mitigated by increasing the number of observations.

\begin{figure*}[ht]
    \centering
    \begin{minipage}[c]{0.155\textwidth} 
        \centering
        \includegraphics[width=0.9\linewidth]{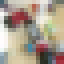}\\
    \end{minipage}
    \vspace{0.5em}
    \begin{minipage}[c]{0.155\textwidth}
        \centering
        \includegraphics[width=0.9\linewidth]{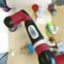}\\
    \end{minipage}
    \vspace{0.5em}
    \begin{minipage}[c]{0.155\textwidth}
        \centering
        \includegraphics[width=0.9\linewidth]{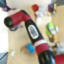}\\
    \end{minipage}
    \vspace{0.5em}
    \begin{minipage}[c]{0.155\textwidth}
        \centering
        \includegraphics[width=0.9\linewidth]{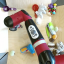}\\
    \end{minipage}
    \vspace{0.5em}
    \begin{minipage}[c]{0.155\textwidth}
        \centering
        \includegraphics[width=0.9\linewidth]{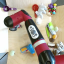}\\
    \end{minipage}

    \vspace{-2em} 

    \begin{minipage}[c]{0.155\textwidth} 
        \centering
        \includegraphics[width=0.9\linewidth]{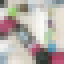}\\
    \end{minipage}
    \vspace{0.5em}
    \begin{minipage}[c]{0.155\textwidth}
        \centering
        \includegraphics[width=0.9\linewidth]{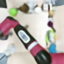}\\
    \end{minipage}
    \vspace{0.5em}
    \begin{minipage}[c]{0.155\textwidth}
        \centering
        \includegraphics[width=0.9\linewidth]{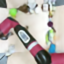}\\
    \end{minipage}
    \vspace{0.5em}
    \begin{minipage}[c]{0.155\textwidth}
        \centering
        \includegraphics[width=0.9\linewidth]{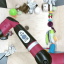}\\
    \end{minipage}
    \vspace{0.5em}
    \begin{minipage}[c]{0.155\textwidth}
        \centering
        \includegraphics[width=0.9\linewidth]{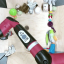}\\
    \end{minipage}

    \vspace{-1.8em} 

    \parbox{0.155\textwidth}{\centering \small \textbf{LR}}%
    \vspace{0.5em}
    \parbox{0.155\textwidth}{\centering \small \textbf{IART \cite{xu2024enhancing}}}%
    \vspace{0.5em}
    \parbox{0.155\textwidth}{\centering \small \textbf{FMA-Net \cite{youk2024fma}}}%
    \vspace{0.5em}
    \parbox{0.155\textwidth}{\centering \small \textbf{VDM-VSR}}%
    \vspace{0.5em}
    \parbox{0.155\textwidth}{\centering \small \textbf{GT}}%
    
    \vspace{-2em} 
    
    \caption{Comparison of 4x super-resolution on BAIR dataset. Only the 5th (middle) frame is shown. }
    \label{fig:bair_comparison}
    \vspace{-0.8em} 
\end{figure*}

\begin{figure*}[t]
    \centering
    
    \begin{minipage}[c]{0.05\textwidth}
        \centering
        \rotatebox{90}{\textbf{LR}}
    \end{minipage}
    \begin{minipage}[c]{0.9\textwidth}
        \centering
        \includegraphics[width=0.16\textwidth]{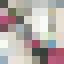}\hspace{0.015\textwidth}
        \includegraphics[width=0.16\textwidth]{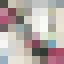}\hspace{0.015\textwidth}
        \includegraphics[width=0.16\textwidth]{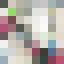}\hspace{0.015\textwidth}
        \includegraphics[width=0.16\textwidth]{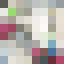}\hspace{0.015\textwidth}
        \includegraphics[width=0.16\textwidth]{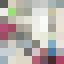}
    \end{minipage}
    
    \begin{minipage}[c]{0.05\textwidth}
        \centering
        \rotatebox{90}{\textbf{Output}}
    \end{minipage}
    \begin{minipage}[c]{0.9\textwidth}
        \centering
        \includegraphics[width=0.16\textwidth]{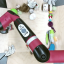}\hspace{0.015\textwidth}
        \includegraphics[width=0.16\textwidth]{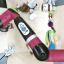}\hspace{0.015\textwidth}
        \includegraphics[width=0.16\textwidth]{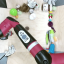}\hspace{0.015\textwidth}
        \includegraphics[width=0.16\textwidth]{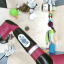}\hspace{0.015\textwidth}
        \includegraphics[width=0.16\textwidth]{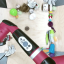}
    \end{minipage}

    \begin{minipage}[c]{0.05\textwidth}
        \centering
        \rotatebox{90}{\textbf{GT}}
    \end{minipage}
    \begin{minipage}[c]{0.9\textwidth}
        \centering
        \includegraphics[width=0.16\textwidth]{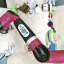}\hspace{0.015\textwidth}
        \includegraphics[width=0.16\textwidth]{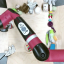}\hspace{0.015\textwidth}
        \includegraphics[width=0.16\textwidth]{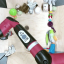}\hspace{0.015\textwidth}
        \includegraphics[width=0.16\textwidth]{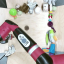}\hspace{0.015\textwidth}
        \includegraphics[width=0.16\textwidth]{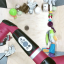}
    \end{minipage}
    
    \caption{Results of our method on BAIR dataset for 8x super-resolution. Only the 1st, 3rd, 5th, 7th, and 9th frame of videos are illustrated.}
    \label{fig:bair_comparison_8x}
    \vspace{-1.8em} 
\end{figure*}

\subsection{BAIR Dataset}
To evaluate our method on real-world data, we used the BAIR robot pushing dataset \cite{ebert2017self}, which consists of 90K short video clips recorded by a real camera. Although this setting remains somewhat of a “toy world” (featuring robotic arms in a controlled environment), it introduces more natural lighting, scene textures, and frequent occlusions than Moving MNIST. We trained our model using 130K video clips, each video contains 10 frames. Our approach effectively reconstructed sharp videos under these conditions, often achieving near-perfect restoration of the robot arm’s position and scene details for both 4x and 8x down-sampled scenarios (see Fig.~\ref{fig:bair_comparison} and Fig.~\ref{fig:bair_comparison_8x}).

We compared our algorithm to two state-of-the-art video super-resolution methods: IART \cite{xu2024enhancing} and FMA-Net \cite{youk2024fma}. For a fair comparison, we retrained these methods on the BAIR robot pushing dataset following their official implementation. We set the down-sampling factor to 4x and measured PSNR and SSIM against the corresponding ground-truth videos for quantitative evaluation (see Table~\ref{tab:psnr_ssim}). Both methods under comparison are based on motion estimation modules, thus presenting limitations in effectively recovering complex motion dynamics. Our method's results highlight that the utilization of inter-frame motion cues improves VSR performance while obviating the requirement for explicit motion estimation.

\begin{table}[ht]
    \centering
    \caption{Quantitative Comparison on BAIR dataset.}
    \label{tab:psnr_ssim}
    \begin{tabular}{l cc cc}
        \toprule
        \textbf{Method} & \textbf{PSNR} & \textbf{SSIM}\\
        \midrule
        IART \cite{xu2024enhancing} & 24.80 & 0.886 \\
        FMA-Net \cite{youk2024fma} & 25.35 & 0.897 \\
        \textbf{VDM-VSR} & \textbf{27.44} & \textbf{0.928} \\
        \bottomrule
    \end{tabular}
    \vspace{-2em} 
\end{table}

\section{Conclusions}
\label{sec:conclude}
In this paper, we introduce a novel VSR algorithm based on the DPS framework, incorporating an unconditional video diffusion model. Unlike the original DPS approach, its reverse diffusion iteration operates in a latent image space extended with a temporal axis, and the denoiser is powered by a transformer neural network. This transformer is pre-trained through an unconditional video diffusion process, enabling it to learn the physics of the world across both spatial and temporal dimensions. We also refined the degradation formula by integrating per-frame downsampling and frame masking to effectively address video super-resolution challenges.

Experiments with synthetic and real-world data revealed that although our algorithm lacks an explicit motion estimation step, it can automatically capture the learned motion patterns from its input and estimate the underlying HR video. Its effectiveness improves with the number of frames used, regardless of the extent of aliasing in the LR videos.

Although these strengths are notable, our current setup cannot yet serve as a fully general-purpose solution, primarily due to limited computational resources and training data. An effective real-world implementation would require a large-scale diffusion model, comparable to those deployed in commercial systems such as OpenAI’s Sora. Nevertheless, our results underscore the capability of advanced video diffusion models to enhance video super-resolution, thereby highlighting a promising direction for both academic research and industrial applications.



\bibliographystyle{IEEEtran}
\bibliography{ref}

\end{document}